# Optimized scheduling of electricity-heat cooperative system considering wind energy consumption and peak shaving and valley filling

Jin Ye
School of Computer and Information Technology
Shanxi University
Taiyuan, China
yejin@just.edu.cn

Lingmei Wang
School of Automation and Software Engineering
Shanxi University
Taiyuan, China
13546468676@163.com

Shujian Zhan
Zhejiang Qingfeng Environmental Co., Ltd.
Lishui, China
kingfit30@qfzl.com

Haihang Wu*
School of Energy and Environment
Southeast University
Nanjing, China
hthh1211@sina.com

*Abstract*— With the global energy transition and rapid development of renewable energy, the scheduling optimization challenge for combined power-heat systems under new energy integration and multiple uncertainties has become increasingly prominent. Addressing this challenge, this study proposes an intelligent scheduling method based on the improved Dual-Delay Deep Deterministic Policy Gradient (PVTD3) algorithm. System optimization is achieved by introducing a penalty term for grid power purchase variations. Simulation results demonstrate that under three typical scenarios (10%, 20%, and 30% renewable penetration), the PVTD3 algorithm reduces the system's comprehensive cost by 6.93%, 12.68%, and 13.59% respectively compared to the traditional TD3 algorithm. Concurrently, it reduces the average fluctuation amplitude of grid power purchases by 12.8%. Regarding energy storage management, the PVTD3 algorithm reduces the end-time state values of low-temperature thermal storage tanks by 7.67–17.67 units while maintaining high-temperature tanks within the 3.59–4.25 safety operating range. Multi-scenario comparative validation demonstrates that the proposed algorithm not only excels in economic efficiency and grid stability but also exhibits superior sustainable scheduling capabilities in energy storage device management.

## I. INTRODUCTION

Energy is the cornerstone of human societal development[1], yet the current fossil fuel-dominated energy structure faces severe challenges in environmental pollution and sustainable development. As the global energy transition accelerates, renewable energy sources with clean and environmentally friendly characteristics are progressively becoming integral components of the energy system[2]. Against this backdrop, integrated energy systems are regarded as key solutions for enhancing renewable energy integration efficiency[3][4], owing to their ability to coordinate and optimize the conversion and storage of multiple energy forms such as electricity, heat, and gas. However, the inherent volatility and intermittency of renewable energy[5], coupled with the complex coupling relationships within electrothermal systems, introduce significant uncertainties into system dispatch. Although the introduction of energy storage technologies provides crucial flexibility for system regulation[6], traditional scheduling methods still exhibit clear limitations in handling multi-source uncertainties. Therefore, researching intelligent scheduling strategies adaptable to high penetration of renewable energy holds significant theoretical and practical importance for ensuring safe and economical system operation while enhancing energy utilization efficiency.

Currently, there have been numerous studies focusing on the field of optimization of energy management strategies. Among them, the optimal energy scheduling techniques based on optimization algorithms have been developed to a mature stage. Literature [7] designed the CSPO-GE optimization algorithm to minimize the economic cost, minimize the carbon emission, and maximize the comprehensive energy efficiency for an integrated hydrogen storage energy system with electric-thermal coupling. On the other hand, literature [8] constructed a two-tier optimization framework for multi-community integrated energy systems, which takes hydraulic stability and economy as the optimization objectives and aims to improve the overall economic efficiency of multi-community integrated energy systems. In addition, literature [9] proposed a two-stage multi-objective benefit equilibrium optimization and coordination strategy for electric-thermal coupled integrated energy systems based on the improved non-dominated sorting genetic algorithm NSGA-II. However, the above research methods are mainly limited to the exploration of the system optimal scheduling problem for fixed scenarios, or the consideration of uncertainty is relatively one-sided, failing to comprehensively and deeply explore various potential uncertainties that may be encountered during system operation [10][11]. In addition, these approaches often face challenges such as high solution complexity, long computation time, and even solution difficulties when dealing with complex systems [12]. In view of these problems, the reinforcement learning-based algorithms have emerged to provide new ideas and ways to solve the above challenges.

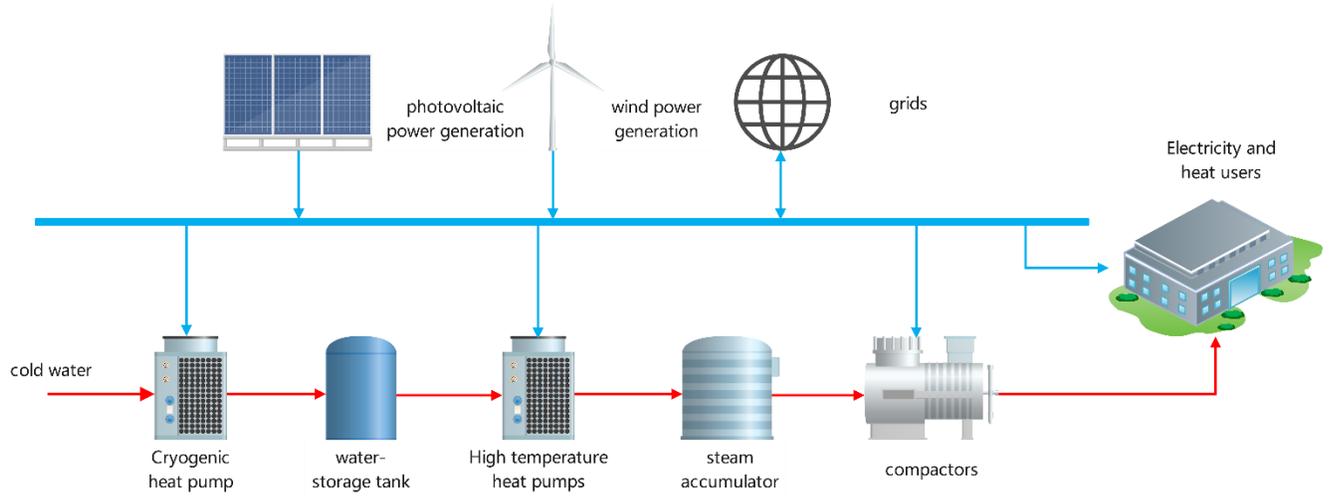

Figure 1 : Electrothermal synergy system diagram

In summary, compared with the traditional optimization-based algorithms, the reinforcement learning-based algorithms show more significant advantages in dealing with uncertainties. In view of this, this paper adopts a model-free reinforcement learning algorithm, the TD3 algorithm, to optimize the scheduling strategy of the electro-thermal cooperative system. The main work of this paper is summarized as follows:

1) A Markov Decision Process (MDP) model for an electric heating system incorporating a steam accumulator energy storage mechanism has been constructed.

2) The improved PVTD3 algorithm is proposed. By incorporating a penalty term for grid power purchase variations into the reward function, the algorithm significantly enhances grid stability while maintaining economic efficiency.

3) A multi-scenario comparative experiment was designed, demonstrating that the proposed algorithm exhibits significant advantages in optimization performance, computational efficiency, and robustness, thereby providing a novel solution for electric heating system scheduling.

II. SYSTEM MODELING AND OPTIMIZATION PROBLEMS

Figure 1 shows the diagram of the electric-thermal co-system with steam accumulator involved in the study, where the whole system consists of two heat pumps, a water storage tank, a steam accumulator, a compactor, photovoltaic wind power generation, an electric grid, and an electric-thermal load. The system not only considers the incorporation of renewable energy sources such as wind power, but also involves the co-dispatch of the grid.

A. Renewable energy model

Renewable power generation can increase the energy supply capacity and reduces the load pressure on the conventional power grid. Therefore, photovoltaic power and wind power are incorporated into the system. The PV power output can be expressed as follows [13]:

Identify applicable funding agency here. If none, delete this text box.

$$P_E^{PV} = \eta_{PV}\eta_{inv}A_{PV}I^0 \quad (1)$$

where $\eta_{PV}$ and $\eta_{inv}$ respectively for the photovoltaic efficiency and inverter efficiency, $I^0$ for the intensity of sunlight radiation, the unit is watts per square meter (W/m2), $A_{PV}$ for the effective light area.

The amount of electricity generated by a wind turbine can be expressed as follows [14]:

$$P_E^{wt} = \begin{cases} 0, & 0 \leq v_t < v_{in},\ v_t > v_{out} \\ \dfrac{v_t - v_{in}}{v_r - v_{in}}Cap_{wt}, & v_{in} \leq v_t \leq v_r \\ Cap_{wt}, & v_r \leq v_t \leq v_{out} \end{cases} \quad (2)$$

where $v_t$, $v_{in}$, $v_{out}$ and $v_r$ are the real-time wind speed, start-up wind speed, shutdown wind speed and rated wind speed, respectively, $Cap_{wt}$ is the installed capacity of the wind turbine, and $P_E^{wt}$ is the power generation capacity of the wind turbine.

B. Heat pump model

The heat pump used in this study is a compression heat pump, which generates heat by consuming electrical energy and its heat generation can be expressed as:

$$P_H^{HP} = COP \cdot P_E^{HP} \quad (3)$$

where $P_E^{HP}$ and $P_H^{HP}$ are the electric power consumed by the heat pump and the thermal power produced, respectively, $COP$ is the heating efficiency of the heat pump.

C. Energy storage equipment model

The energy storage devices in this study involve water storage tanks and steam heat accumulators, which are often used to mitigate the uncertainties of renewable energy sources and customer loads. In this study, the states of the energy storage devices are all expressed using the amount of heat stored, and the change in their heat storage state can be expressed by the following equation:

$$HSD\_L_{t+1} = HSD\_L_t + Q_{L,t}^{in} - Q_{L,t}^{out} \quad (4)$$

$$HSD\_H_{t+1} = HSD\_H_t + Q_{H,t}^{in} - Q_{H,t}^{out} \quad (5)$$

where $HSD\_L_t$ and $HSD\_H_t$ are the energy storage states of the two storage devices at the current moment t, $Q_{L,t}^{in}$ and $Q_{L,t}^{out}$ are the heat entering and leaving the hot water storage tank, $Q_{H,t}^{in}$ and $Q_{H,t}^{out}$ are the heat entering and leaving the steam heat accumulator, $HSD\_L_{t+1}$ and $HSD\_H_{t+1}$ are the states of the storage devices after updating the states, i.e., at the next moment, respectively.

### D. Compactor Model

The high temperature steam from the steam accumulator is compressed by a compactor to obtain superheated steam and used to supply the heat load demand of the user. The mathematical model can be expressed as:

$$P_H^{CP} = P_E^{CP} \cdot \eta_{CP} \quad (6)$$

where $P_E^{CP}$ is the electrical power consumed by the compressor, $\eta_{CP}$ is the compression efficiency of the compressor, $P_H^{CP}$ is the thermal power generated by the compressor.

### E. Optimization problem formulation

During the operation of the system, the water storage tank and the steam accumulator have three states of charging, discharging and not working, respectively. The operating power of the heat pump can be appropriately adjusted according to the current heat load demand and the energy storage state of the water storage tank and steam accumulator. For the scheduling optimization problem of this electro-thermal synergistic system, the power output of the heat pump and compactor needs to be accurately regulated in order to minimize the scheduling cost and enhance the overall sustainability of the system. Specifically, the system scheduling cost mainly comes from the cost of grid power purchase.

To ensure stable and continuous operation of the system, at the end of the daily dispatch cycle, the energy storage status of the water storage tanks and steam accumulator tanks should be restored to the initial level as much as possible to adapt to the demand of the thermo-electrolytic coupling on the following day. The objective function, constraints, and power balance of the optimization problem are described in detail below:

1) Objective function

$$J = \min(C_1 + C_2)$$
$$\begin{cases} C_1 = k_1 \sum_k P_E^G \cdot p \\ C_2 = k_2 \Delta HSD\_L_T + k_3 \Delta HSD\_H_T \\ \quad = k_2(HSD\_L_T - HSD\_L_0) + k_3(HSD\_H_T - HSD\_H_0) \end{cases} \quad (7)$$

where $C_1$ is the cost of purchasing and selling electricity from the grid, $C_2$ is the cost of the terminal storage state of the water storage tanks and steam accumulator tanks, $HSD\_L_T$ and $HSD\_H_T$ are the storage states of the equipment at the end of the scheduling cycle of the water storage tanks and steam accumulator tanks, respectively, $HSD\_L_0$ and $HSD\_H_0$ are the initial storage states of the corresponding equipment. $k_1$, $k_2$ and $k_3$ are empirically adjusted constant factors.

2) restrictive condition

$$P_{E,k}^{PV} + P_{E,k}^{Wind} + P_{E,k}^{G} = P_{E,k}^{HP\_L} + P_{E,k}^{HP\_H} + P_{E,k}^{L} + P_{E,k}^{CP} \quad (8)$$

$$P_{H,k}^{CP} = P_{H,k}^{L} \quad (9)$$

$$HSD\_L_{\min} \leq HSD\_L_k \leq HSD\_L_{\max} \quad (10)$$

$$HSD\_H_{\min} \leq HSD\_H_k \leq HSD\_H_{\max} \quad (11)$$

$$P_{E,\min}^{HP\_L} \leq P_{E,k}^{HP\_L} \leq P_{E,\max}^{HP\_L} \quad (12)$$

$$P_{E,\min}^{HP\_H} \leq P_{E,k}^{HP\_H} \leq P_{E,\max}^{HP\_H} \quad (13)$$

$$P_{E,\min}^{CP} \leq P_{E,k}^{CP} \leq P_{E,\max}^{CP} \quad (14)$$

The problem involves the optimization of a cumulative objective function with multiple decision variables, which needs to overcome various state and control constraints at each point in time, while these variables interact with each other in time and state space, making the problem complex and challenging to solve directly. In addition, there are multiple uncertainties in the system, such as fluctuations in the output of photovoltaic and wind power, as well as variations in the electrical and thermal loads, which require accurate prediction models and cumbersome computational processes. In view of this, the TD3 algorithm, which does not require a prediction model, is used in this paper to achieve efficient real-time energy scheduling of the electro-thermal cooperative system and to enhance the adaptability of the system.

## III. TWIN DELAYED DEEP DETERMINISTIC POLICY GRADIENT

### A. Markov decision process

According to the operational characteristics of the system, the following are the specific definitions of the state space, action space and reward function:

1) State space

$$\boldsymbol{S} = \left[ t, P_E^L, P_H^L, P_E^{PV}, P_E^{Wind}, HSD\_L, HSD\_H \right]^T \quad (15)$$

where t is the time step, $P_E^L$ and $P_H^L$ are the electric and thermal loads at the current moment, $P_E^{PV}$ and $P_E^{Wind}$ are the photovoltaic and wind power generation, $HSD\_L$ and $HSD\_H$ are the energy storage status of the water and steam heat storage tanks, respectively.

2) action space

$$\boldsymbol{A} = \left[ P_E^{HP\_L}, P_E^{HP\_H}, P_E^{CP} \right]^T \quad (16)$$

where $P_E^{HP\_L}$, $P_E^{HP\_L}$, and $P_E^{CP}$ denote the power consumption of low-temperature heat pumps, high-temperature heat pumps, and compressors, respectively.

3) Reward function

$$R = -C_1 - C_2 - l_1 P_{ex}^E - l_2 P_{loss}^E - l_3 P_{ex}^H - l_4 P_{loss}^H - l_5 \Delta P_t \quad (17)$$

where the first half of the definition of $R$ is consistent with the objective function and contains the cost of power purchase and sale from the grid and the state cost of the energy storage device. In the initial training stage, although the actions output by the actor network have taken into account the charging and discharging operations of the energy storage device, it may still not strictly guarantee the balance between power supply and demand. Therefore, this study introduces an additional penalty term mechanism in the reward function, aiming to enhance the energy utilization efficiency of the system. Specifically, by adding penalty terms for deviations in the supply and demand of electrical and thermal energy, the system can effectively optimize the energy allocation strategy, thereby avoiding energy waste and ensuring user demand is met. Additionally, to promote the peak shaving and valley filling effect of the grid, this study further incorporates a penalty term for fluctuations in the grid's power purchase in the reward function, to smooth out power fluctuations during the dispatching process. The introduction of these penalty terms will significantly accelerate the strategy optimization of the actor network during the training process, thereby improving the overall operational efficiency and energy management performance of the electro-thermal integrated system.

## B. TD3 training parameters and network initialization

The selection of TD3 training parameters is decisive for the performance of the algorithm and the performance of the results. Based on the above considerations, the training parameters set in this paper are shown below:

Table 1 Main parameters of TD3

| parameters | value | parameters | value |
|---|---|---|---|
| $a^P$ | 0.0001 | $\tau$ | 0.01 |
| $a^Q$ | 0.001 | $k_1$ | 0.1 |
| episode | 200 | $k_2$ | 0.055 |
| $step_{delay}$ | 3 | $k_3$ | 0.1 |
| $\gamma$ | 0.99 | $l_1$ | 1.1 |
| memory | 4000 | $l_2$ | 1 |
| Batch size | 64 | $l_3$ | 1.2 |
| $\mu$ | 0 | $l_4$ | 1 |
| $\theta$ | 0.15 | $l_5$ | 1.5 |

## IV. SIMULATION RESULTS AND ANALYSIS

### A. Typical Day Simulation Scenario

Before the start of each training round, the system randomly generates a set of system operating conditions based on typical day scenarios and feasible domains for safe equipment operation. These working conditions include parameters such as solar irradiance, wind speed, electrical load and thermal load to simulate multiple uncertainties in real operation. Fig. 3 shows the variation of electrical load, thermal load, PV output and wind power generation for a typical day scenario.

In order to simulate the uncertainty in actual operation, this study introduces a volatility of no more than 30% based on the typical day scenario as a way to simulate the uncertainty situation that the system may encounter. Such a treatment can more realistically assess the performance and stability of the system when facing load variations. Figure 2 illustrates the range of fluctuation of the load data in the training scenario, which is also the range of data on which the algorithm is trained. Specifically, the range of load data fluctuations in Figure 2 reflects the potential changes in the system load under different levels of uncertainty, thus providing the algorithm with a variety of training scenarios to ensure that it maintains good adaptability and predictive capability when facing various uncertainties in real operations.

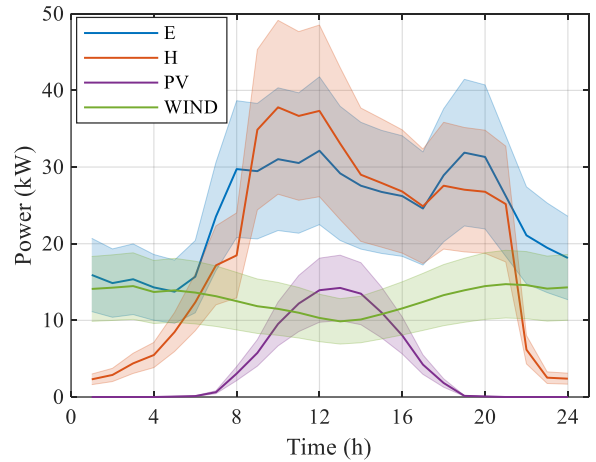

Figure 2: Typical day scenario load

### B. Typical Day Scenario Scheduling Results

Figure 3 shows the training state of the TD3 algorithm, where "mean" denotes the average of the cumulative rewards in every 5 training cycles. The training curve shows that at the beginning of the training period, due to the introduction of noise, the cumulative reward of the TD3 algorithm is low, and reaches a minimum of about -2200 at about 5 training rounds, and the cumulative reward of the TD3 algorithm gradually improves as the training process continues, and begins to converge at about 150 training rounds. Eventually, the cumulative reward of the algorithm stabilizes at about -200, indicating that the algorithm has successfully learned an effective policy and has stable performance.

Figure 4 illustrates the change in energy storage state of the water storage tank and the steam accumulator for a typical day scenario. For clarity of representation, the energy storage states of both the water storage tank and the steam accumulator are expressed in kWh. The upper part of the figure shows the change of the energy storage state of the water storage tank, and it can be seen that in the first 7 moments, the decision intelligence tends to fill the water storage tank for subsequent hot water usage demand, and gradually returns to the initial state at the end of the dispatch cycle. The lower part of the figure shows the change in the energy storage state of the steam accumulator tank, and it is clear that in the initial moments, the decision intelligence replenishes the energy of the steam accumulator tank for subsequent use, and gradually returns to the initial state at the end of the scheduling cycle. Overall, both the water storage tank and the steam accumulator tank tend to replenish energy at the beginning of the scheduling period to meet the subsequent heating demand.

Figure 5 illustrates the power balance for a typical day scenario. It can be clearly seen from the figure that in the first seven moments of the scheduling cycle, the system utilizes the low tariff period to purchase a large amount of electricity and uses the wind-generated electricity for water storage in the water storage tanks. With the increase of electric load demand during the daytime, it is difficult for wind power generation to meet the demand for electricity, so the system therefore purchases a small amount of electricity from the grid to meet the user demand, and at the same time controls the water storage tanks not to be fed with water anymore to minimize the unnecessary loss of purchased electricity. In addition, the system fills the steam accumulator with energy to cope with the increasing thermal load demand. At the end of the dispatch cycle, the system gradually reduces the energy charge to the steam accumulator to restore its state to the initial moment of dispatch.

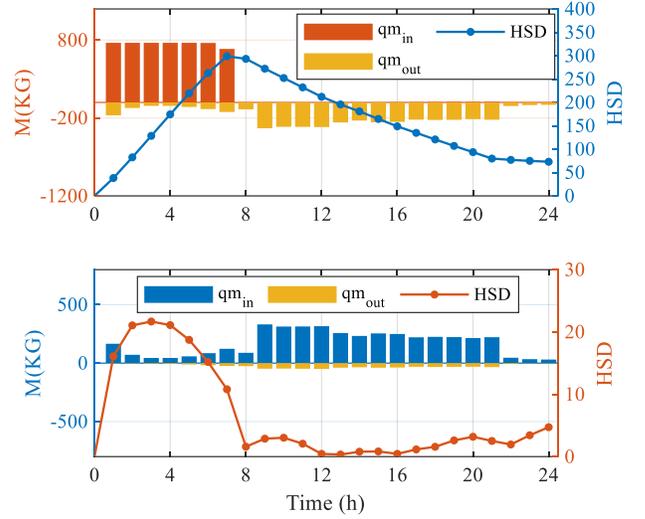

Figure 4: state changing of water and steam heat storage tanks

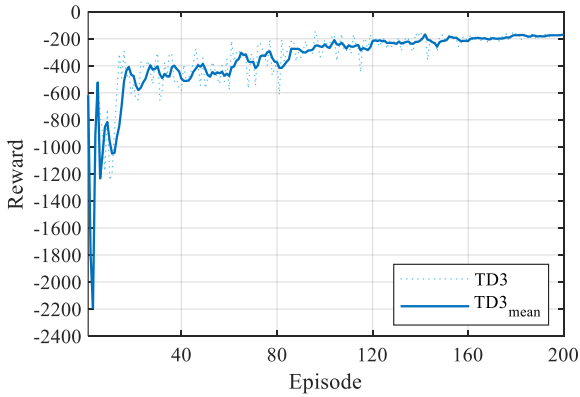

Figure 3: Training Curve

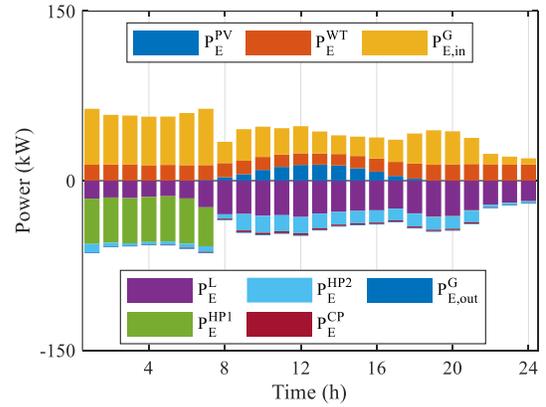

Figure 5: Power balance diagram for typical day scenario

C. *Comparative analysis of algorithms in multiple scenarios*

To validate the superiority and robustness of the algorithms under various uncertainty conditions, this study compares and analyzes the TD3 algorithm and the peak-shaving and valley-filling TD3 algorithm (PVTD3) in three environments with different uncertainty levels. To validate the superiority and robustness of these algorithms under multiple uncertainty conditions, this study compares and analyzes the TD3 algorithm and the PRTD3 algorithm across three levels of uncertainty. Specifically, a total of 30 random scenarios were selected from uncertainty levels of 0-10%, 10%-20%, and 20%-30% as comparison scenarios to evaluate the performance of different algorithms in varying uncertainty environments. Specifically, a total of 30 random scenarios were selected from uncertainty levels of 0-10%, 10%-20%, and 20%-30% as comparison scenarios to evaluate the performance of different algorithms under varying uncertainty conditions.

In Table 2, $C\_s$ represents the integrated cost of dispatch accumulated by the system over a complete dispatch cycle. This cost encompasses the electricity procurement expenses for the system during the cycle and the operational maintenance costs for the energy storage facilities. It should be noted that the figures presented in the table represent the average calculated across ten distinct uncertainty scenarios.

Table 2 compares the scheduling performance of the two algorithms in different scenarios (10%, 20%, 30%) for the electric-thermal coordination system. The data indicates that the PVTD3 algorithm outperforms the traditional TD3 algorithm in both comprehensive cost and grid power purchase variation. Specifically, under the 10%, 20%, and 30% scenarios, PVTD3 reduced the comprehensive cost by 6.93%, 12.68%, and 13.59% respectively compared to TD3, demonstrating superior economic optimization capabilities. Concurrently, the grid power purchase variation under PVTD3 decreased by approximately 12.8% on average compared to TD3, indicating enhanced performance in mitigating grid power fluctuations.

Table 2 Comparison of algorithm scheduling results under different uncertainty scenarios

|  | TD3 | | | PVTD3 | | |
| --- | --- | --- | --- | --- | --- | --- |
|  | 10% | 20% | 30% | 10% | 20% | 30% |
| $HSD\_L_T$ | 80.78 | 79.66 | 83.26 | **73.05** | **70.07** | **65.59** |
| $HSD\_H_T$ | 4.76 | 7.06 | 7.49 | **4.24** | **4.25** | **3.59** |
| C_s/万元 | 1014.1 | 1052.1 | 1151.1 | **943.81** | **918.69** | **994.85** |
| $\Delta P$ | 303.68 | 306.67 | 297.70 | **264.82** | **268.57** | **255.40** |

## V. Conclusion

This study addresses the challenges of multi-source uncertainties, renewable energy integration, and grid peak-shaving demands in the operation of combined electric-thermal systems. An integrated optimization model incorporating a steam thermal storage mechanism was developed, alongside an intelligent scheduling solution based on an improved TD3 algorithm. Simulation validation yielded the following key conclusions:

1) The Markov Decision Process (MDP) model constructed in this paper is designed to accurately characterize the dynamic operation of the system by fully taking into account the multidimensional uncertainties involved in the system operation, effectively coping with the challenges of consumption in the integration of new energy sources, and taking into account the peak shaving and valley filling needs in the power demand management.
2) The comprehensive cost of system was reduced by 6.93%, 12.68%, and 13.59% under three typical scenarios of 10%, 20%, and 30% respectively, demonstrating the PVTD3 algorithm's significant economic optimization capability.
3) The averaged fluctuation range of grid power purchase under the PVTD3 algorithm decreased by 12.8%, effectively achieving the peak-shaving and valley-filling dispatch objective.

## References

[1] Stennikov V, Barakhtenko E, Sokolov D, et al. Current state of research on the energy management and expansion planning of integrated energy systems[J]. Energy Reports, 2022, 8: 10025-10036.
[2] Mohseni S, Brent A C, Kelly S, et al. Demand response-integrated investment and operational planning of renewable and sustainable energy systems considering forecast uncertainties: A systematic review[J]. Renewable and Sustainable Energy Reviews, 2022, 158: 112095.
[3] Xuan A, Shen X, Guo Q, et al. Two-stage planning for electricity-gas coupled integrated energy system with carbon capture, utilization, and storage considering carbon tax and price uncertainties[J]. IEEE Transactions on Power Systems, 2022, 38(3): 2553-2565.
[4] Ran J, Qiu Y, Liu J, et al. Coordinated optimization design of buildings and regional integrated energy systems based on load prediction in future climate conditions[J]. Applied Thermal Engineering, 2024, 241: 122338.
[5] Igeland P, Schroeder L, Yahya M, et al. The energy transition: The behavior of renewable energy stock during the times of energy security uncertainty[J]. Renewable Energy, 2024, 221: 119746.
[6] Azarhooshang A, Sedighizadeh D, Sedighizadeh M. Two-stage stochastic operation considering day-ahead and real-time scheduling of microgrids with high renewable energy sources and electric vehicles based on multi-layer energy management system[J]. Electric Power Systems Research, 2021, 201: 107527.
[7] Zhong S, Wang X, Wu H, et al. Energy hub management for integrated energy systems: A multi-objective optimization control strategy based on distributed output and energy conversion characteristics[J]. Energy, 2024, 306: 132526.
[8] Chen Z, Zhao W, Lin X, et al. Load prediction of integrated energy systems for energy saving and carbon emission based on novel multi-scale fusion convolutional neural network[J]. Energy, 2024, 290: 130181.
[9] Zhang N, Feng T. Two-stage Multi-objective Optimization Coordination of Electro-thermal Coupled Integrated Energy System Based on Improved NSGA-II Algorithm[J]. Distributed Generation & Alternative Energy Journal, 2023: 1707-1740.
[10] Wang Z, Liu Z, Huo Y. A distributionally robust optimization approach of multi-park integrated energy systems considering shared energy storage and Uncertainty of Demand Response[J]. Electric Power Systems Research, 2025, 238: 111116.
[11] Gao J, Meng Q, Liu J, et al. Thermoelectric optimization of integrated energy system considering wind-photovoltaic uncertainty, two-stage power-to-gas and ladder-type carbon trading[J]. Renewable Energy, 2024, 221: 119806.
[12] Dong Y, Zhang H, Wang C, et al. Soft actor-critic DRL algorithm for interval optimal dispatch of integrated energy systems with uncertainty in demand response and renewable energy[J]. Engineering Applications of Artificial Intelligence, 2024, 127: 107230.
[13] Zhou Y, Ma Z, Shi X, et al. Multi-agent optimal scheduling for integrated energy system considering the global carbon emission constraint[J]. Energy, 2024, 288: 129732.
[14] Keyhani A, Ghasemi-Varnamkhasti M, Khanali M, et al. An assessment of wind energy potential as a power generation source in the capital of Iran, Tehran[J]. Energy, 2010, 35(1): 188-201.